\documentclass[letterpaper]{article} 
\usepackage{aaai2027}  
\usepackage[hyphens]{url}  
\usepackage{graphicx} 
\usepackage{natbib}  
\usepackage{caption} 
\usepackage{algorithm}
\usepackage{algorithmic}
\usepackage{amsmath}
\usepackage{amssymb}
\usepackage{multirow}

\usepackage{newfloat}
\usepackage{listings}
\DeclareCaptionStyle{ruled}{labelfont=normalfont,labelsep=colon,strut=off} 
\floatstyle{ruled}
\newfloat{listing}{tb}{lst}{}
\floatname{listing}{Listing}

\usepackage{booktabs}

\title{LookME: Lookup-Based Multimodal Embeddings for Layer Injection in Vision-Language Models}
\author{
    Zeyu Xu\equalcontrib,
    Xingzhong Hou\equalcontrib,
    Pengkai Guo,
    Siling Lin,
    Xiao Xu, \\
    Menghua Zhai,
    Haoyu Chen,
    Yunke Zhang\corresponding,
    Fei Huang\corresponding
}
\affiliations{
    Honor Device Co., Ltd\\
    zhangyunke@honor.com, huangbo6@honor.com
}

\begin{document}

\maketitle

\begin{abstract}
Vision-Language Models (VLMs) have achieved strong progress in multimodal understanding. However, scaling dense or sparse Mixture-of-Experts (MoE) models to improve performance limits deployment in resource-constrained environments due to the trade-off between high memory usage from full loading and increased latency from on-demand loading. Recently, the Per-Layer Embedding (PLE) architecture addresses this  by scaling models with large external embedding tables stored in read-only memory (ROM) and performing lightweight lookup to retrieve relevant embeddings to enhance token representations. Nevertheless, existing PLE-style methods are primarily designed for text embeddings  due to the convenience of ID-based retrieval, limiting their effectiveness in VLMs where multimodal embeddings contain richer information for visual tasks. In this paper, we propose LookME, the first framework that enables lookup-based enhancement for multimodal embeddings in VLMs while supporting partitioned storage and on-demand loading. To efficiently lookup arbitrary continuous multimodal embeddings from large-scale embedding tables, we propose a hierarchical two-level lookup method employing a coarse-to-fine strategy that performs lookups from the scene-level to the intra-scene primitive-level. Furthermore, we integrate the lookup method with a sparse injection strategy, which adaptively prioritizes critical embeddings over voluminous multimodal embeddings within layers, and facilitates embedding table reuse across neighboring layers, improving the trade-off among efficiency, model size, and performance. Experiments on multiple visual benchmarks show that LookME outperforms text-only PLE-style methods, validating the effectiveness of lookup-based multimodal embedding enhancement.
\end{abstract}


\section{Introduction}
Vision-Language Models (VLMs) have become a dominant paradigm for multimodal intelligence, achieving remarkable performance across diverse visual tasks~\cite{liu2023llava}. Following scaling laws~\cite{kaplan2020scalinglawsneurallanguage}, recent VLMs have expanded to 72B parameters or larger~\cite{bai2025qwen25vltechnicalreport,chen2024internvl}. However, dense scaling incurs prohibitive FLOPs, bandwidth, and energy costs~\cite{hoffmann2022chinchilla}. Mixture-of-Experts (MoE) methods activate only a fraction of parameters via token routing~\cite{shazeer2017moe,fedus2022switch}, enabling larger scales with low computational cost. Nevertheless, MoE typically requires sufficient GPU memory to store the entire model. Otherwise, frequent loading of large, non-contiguous expert weights introduces significant latency, hindering deployment on resource-constrained edge devices~\cite{wu2024deepseekvl2mixtureofexpertsvisionlanguagemodels}.

Recently, Per-Layer Embedding (PLE) has been proposed as an alternative paradigm for sparse parameter activation~\cite{google2025gemma3n}. Instead of directly increasing the number of model parameters, PLE augments the model with large-scale token-level external knowledge embedding tables. During inference, PLE performs lightweight lookups to integrate relevant embeddings for information enhancement. By storing knowledge tables on offline media (e.g., ROM)~\cite{ding2026mekimemorybasedexpertknowledge} rather than GPU memory, PLE expands model capacity while remaining ideal for resource-constrained edge devices. However, existing PLE methods are mainly designed for LLMs~\cite{cheng2026conditionalmemoryscalablelookup,stem2026,liu2026scalingembeddingsoutperformsscaling}, where discrete text tokens naturally enable efficient lookup. Although some multimodal models adopt PLE~\cite{google2025gemma3n}, the enhancement is largely limited to text tokens, making it insufficient for VLMs dominated by multimodal tokens~\cite{dosovitskiy2021vit}. Extending PLE to VLMs thus requires efficient retrieval over continuous multimodal embeddings. Existing methods, including similarity-based nearest-neighbor retrieval and classification-based large-vocabulary routing, still match each token against the full table and assume the embedding table stays online, requiring joint loading with the model and preventing the decoupled storage and efficient retrieval required by PLE.

In this paper, we propose LookME, the first PLE-style framework enabling lookup-based enhancement for multimodal embeddings in VLMs. As illustrated in Figure~\ref{fig:fig_intro}, LookME supports partitioned storage and on-demand loading, extending lookup mechanisms beyond text to the dominant multimodal stream. Since multimodal embeddings lack inherent discrete IDs, we introduce a hierarchical two-level lookup method, which first coarsely routes embeddings to individual scenes, then performs fine-grained retrieval based on the relevance to scene-primitives. This design ensures balanced lookups while preventing collapse for a large-scale embedding table, naturally extending to all multimodal embeddings like multi-scale image embeddings and cross-modal embeddings. Furthermore, to address redundancy and parameter overhead, we propose a sparse injection strategy that adaptively selects critical embeddings for lookup within layers and dynamically reuses tables across layers, enhancing both efficiency and performance. Experiments on multiple benchmarks demonstrate that LookME consistently outperforms baselines and text-only PLE-style methods.
Our main contributions can be summarized as:
\begin{figure}[t]
\centering
\includegraphics[width=0.98\columnwidth]{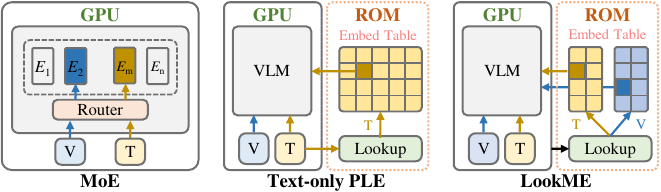} 
\caption{Comparison of architectures between different sparse models.}
\label{fig:fig_intro}
\end{figure}
\begin{itemize}
    \item We propose LookME, a framework that extends PLE-style methods from text to multimodal embeddings, enabling offline embedding construction, lookup, and enhancement for dominant multimodal information in VLMs.
    \item We design a multimodal embedding lookup method that employs ID mapping for discrete text embeddings and a hierarchical two-level lookup method for continuous image and cross-modal embeddings. 
    \item We design a sparse injection strategy that adaptively injects looked-up embeddings within and across layers, striking a balance between efficiency, model size, and performance.
    \item Experiments on multiple visual task benchmarks demonstrate that LookME outperforms baselines as well as text-only PLE-style methods.
\end{itemize}

\section{Related Works}
\subsection{Parameter Scaling Beyond RAM}
Following scaling laws, VLMs have rapidly expanded to 72B+ parameters for stronger multimodal understanding~\cite{bai2025qwen25vltechnicalreport,chen2024internvl,lu2024deepseekvlrealworldvisionlanguageunderstanding}, but this also brings substantial computation and memory overhead. Sparse parameter scaling addresses this issue by increasing model capacity while activating only a subset of parameters. MoE achieves strong performance through dynamic expert routing~\cite{dai2024deepseekmoeultimateexpertspecialization}, yet its need to keep all parameters on-device to avoid switching latency, resulting in high memory costs. To address this challenge, Gemma-3~\cite{google2025gemma3n} proposes PLE, a parameter-efficient lookup embedding method that retrieves ROM-stored embeddings with text IDs for per-token enhancement, balancing memory efficiency and model quality. MeKi~\cite{ding2026mekimemorybasedexpertknowledge} further introduces token-based embeddings with re-parameterization, enabling compact ROM-stored lookup tables that support low-latency loading. Engram~\cite{cheng2026conditionalmemoryscalablelookup} and LongCAT~\cite{liu2026scalingembeddingsoutperformsscaling} extend PLE from single tokens to N-gram retrieval via specialized hashing and contextualized gating, achieving MoE-competitive performance. STEM~\cite{stem2026} further enables efficient sparse scaling by replacing up-projections with token-indexed lookups, improving stability while reducing overhead. However, existing PLE-style methods only support lookup over discrete text embeddings, limiting their applicability to VLMs. LookME extends PLE to VLMs by enabling lookup over multimodal embeddings.

\subsection{Embedding Lookup in VLMs}
Embedding-based lookup in VLMs mainly falls into three categories. \emph{Codebook-based} methods, such as VQ-VAE~\cite{van2017vqvae} and residual/product quantization~\cite{jegou2011product}, map continuous embeddings to discrete codes for compression and reconstruction rather than representation enhancement, making them less suitable for model scaling. 
\emph{Similarity-based} methods retrieve embeddings via dot-product or cosine similarity~\cite{khandelwal2020knnlm} between tokens and an embedding table~\cite{lample2019productkey,berges2024memorylayers} or image tokenizer~\cite{lu2024ovis}. Since each token must be matched against the entire table, retrieval incurs high latency. Although a two-level coarse-to-fine strategy can accelerate retrieval~\cite{gupta2025retreever,gupta2025cobweb}, both stages remain similarity-based and still require loading the full embedding table, effectively making it part of a dense model. This prevents decoupling parameter storage from sparse on-demand activation, limiting practicality for ROM-resident deployment. 
\emph{Classification-based} methods formulate embedding lookup as slot assignment, directly predicting slot distributions with a classifier without loading the embedding table~\cite{shen2023vlmoe}. This design enables fast retrieval, but the lack of table guidance limits retrieval precision and may cause routing collapse. Moreover, the classifier scales with table size, must remain fully online, and incurs substantial computation, limiting efficiency on resource-constrained devices. 
LookME adopts a \emph{hierarchical two-level} design, requiring only a vocabulary-log-scale similarity embedding matrix and classification head to reside on GPU, while storing the full embedding table independently in ROM for fast and efficient retrieval with minimal GPU overhead.

\section{Methods}

\subsection{Overview}
\label{sec:overview}
\begin{figure*}[t]
\centering
\includegraphics[width=0.98\textwidth]{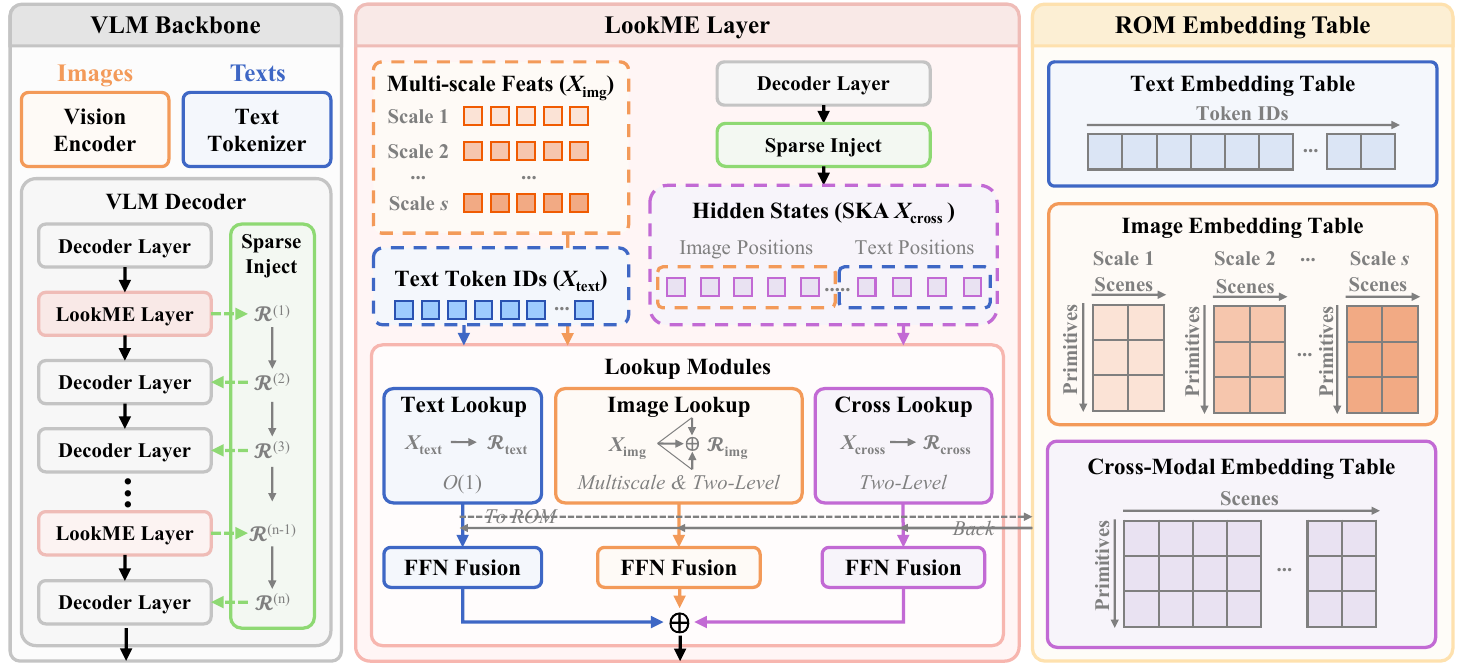}
\caption{Overview of LookME. External embedding tables are stored in ROM, while LookME layers are inserted with sparse intra-layer injection. Each layer uses sparse inter-layer injection and three parallel lookup branches for enhancement.}
\label{fig:fig_method}
\end{figure*}
Text-only PLE methods enhance the hidden states $H_{\text{in}}\in \mathbb{R}^{L \times d_\text{h}}$ at each predefined layer as:  
\begin{equation}
\begin{gathered}
    H_{\text{out}} = \mathcal{L}_{\text{dec}}(H_{\text{in}}) + \mathcal{G}(\mathcal{R}_{\text{text}}), \quad   \mathcal{R}_{\text{text}} = E[u],
\end{gathered}
\end{equation}
where $L$ denotes the sequence length, $d_\text{h}$ denotes the hidden dimension of the decoder layers $\mathcal{L}_{\text{dec}}$, $\mathcal{R}_{\text{text}}$ the retrieved external embeddings using the input text token IDs $u$ from an external embedding table $E \in \mathbb{R}^{|V| \times d_\text{e}}$ with $|V|$  representing the vocabulary size, $d_\text{e}$ the embedding dimension, and $\mathcal{G}$ the injection function. The process leaves the dominant multimodal tokens in VLMs untouched and operates on all positions indiscriminately, regardless of their heterogeneous signal enhancement requirements.

To address these limitations, we propose LookME, as illustrated in Figure~\ref{fig:fig_method}. LookME layers $\mathcal{L}_{\text{aug}}$ are integrated into intermediate decoder layers, supporting \emph{Text}, \emph{Image}, and \emph{Cross-Modal} as three complementary pathways for lookup. By employing sparse injection, $\mathcal{L}_{\text{aug}}$ selectively augments the decoder with retrieved multimodal external embeddings across tokens and layers:
\begin{equation}
\begin{aligned}
\label{eq:layer-output}
    H_{\text{out}} = \mathcal{L}_{\text{aug}}(H_{\text{in}}) = & \mathcal{L}_{\text{dec}}(H_{\text{in}}) \;+\; \\ & \mathcal{S}\!\left(\mathcal{R}_{\text{text}}\right) + \mathcal{S}\!\left(\mathcal{R}_{\text{img}}\right) + \mathcal{S}\!\left(\mathcal{R}_{\text{cross}}\right),
\end{aligned}
\end{equation}
where $\mathcal{R}_{*}$ denotes the per-modality pathway retrieval (in the Multimodal Embedding Lookup Section) and $\mathcal{S}$ denotes the sparse injection (in the Sparse Injection Section).

\subsection{Multimodal Embedding Lookup}
\label{sec:multimodal-lookup}
For a given LookME layer $\mathcal{L}_{\text{aug}}^{(l)}$, the multimodal embedding lookup module retrieves external embeddings through three complementary pathways to enhance $H_{\text{in}}$. Each pathway $\mathcal{R}_{*}$ contains an individual input ${X}_{*}$ and a retrieval function $\mathcal{T}_{*}$ that retrieves embeddings from its external embedding table ${E}_{*}$:
\begin{equation}
\begin{aligned}
    \mathcal{R}_{\text{text}}^{(l)} &= \mathcal{T}_\text{text}(X_\text{text};\ E_{\text{text}}^{(l)}), \\
    \mathcal{R}_{\text{img}}^{(l)} &= \mathcal{T}^{(l)}_\text{img}(X_\text{img};\ E_{\text{img}}^{(l)}), \\
    \mathcal{R}_{\text{cross}}^{(l)} &= \mathcal{T}^{(l)}_\text{cross}(X_\text{cross}^{(l)};\ E_{\text{cross}}^{(l)}).
\end{aligned}
\end{equation}
The external embedding tables ${E}_{*}$ for each pathway are initialized via principal component analysis (PCA) on pretrained backbone embeddings, providing alignment with the backbone model. During training, they are jointly optimized together with the backbone, while at inference they can be decoupled and stored separately in ROM, using the input ${X}_{*}$ and retrieval function $\mathcal{T}_{*}$ for lookup. Details of each pathway are described below.

\subsubsection{Text Embedding Index-Based Lookup}
The text pathway operates on text positions in the image-query input. Following text-only PLE methods, it directly uses the corresponding text token IDs $u$ as the retrieval input ${X}_{\text{text}}$ to retrieve text external embeddings $E_{\text{text}}[u]$. This $O(1)$ lookup injects context-independent lexical priors into the text positions.

\subsubsection{Image Embedding Hierarchical Two-Level Lookup}
\label{sec:image-lookup}
The image pathway operates on image positions in the image-query input. The image embeddings $v \in \mathbb{R}^{L_\text{v} \times d_\text{v}}$ produced by the visual encoder are used as the retrieval input ${X}_{\text{img}}$, where $L_\text{v}$ and $d_\text{v}$ denote the number and dimensionality of image embeddings, respectively. These embeddings provide stable and representative retrieval features without extra computation. Since continuous image embeddings lack discrete vocabulary IDs, retrieval is fundamentally formulated either as similarity matching (via a projection $p \in \mathbb{R}^{d_\text{v} \times d_\text{e}}$ against the embedding table $E_{\text{img}} \in \mathbb{R}^{|V| \times d_\text{e}}$) or classification (via a weight $w \in \mathbb{R}^{d_\text{v} \times |V|}$). As the visual embedding table expands to a text-equivalent scale, computation cost and routing imbalance increase significantly, resulting in massive parameter under-utilization. To overcome this limitation, we design a hierarchical two-level lookup method. Image embeddings are initially routed to a coarse-grained \emph{scene}, followed by a dynamic soft-combination of specific \emph{scene primitives} to enrich details. Details are presented in the following sections.

\begin{itemize}
    \item \textbf{Two-level embedding table.}
    For hierarchical two-level retrieval, we reorganize the image external embedding table into a two-dimensional structure $E_{\text{img}} \in \mathbb{R}^{N_\text{s} \times N_\text{p} \times d_\text{e}}$, 
    with $N_\text{s}$, $N_\text{p}$ denoting the numbers of \emph{scenes} and \emph{scene-primitives}, respectively, and $N_\text{s} \times N_\text{p}$ being comparable to $|V|$.  This factorization provides statistical stability at the scene level and fine-grained expressivity at the scene-primitive level, without increasing the number of parameters compared with a flat table of the same total size.

    \item \textbf{Level-1: scene retrieval.}
    \label{sec:level-1}
    Level-1 performs coarse-grained scene-level retrieval by categorizing image embeddings into different scenes according to their global semantics, thereby facilitating subsequent fine-grained retrieval within the selected scene. We use scene-centered embeddings $e_{\text{img}}\in \mathbb{R}^{N_\text{s} \times d_\text{e}}$ to represent scene features, which are dynamically updated during training:
    \begin{equation}
        e_{\text{img}} = \frac{1}{N_\text{p}} \sum_{i=1}^{N_\text{p}} E_{\text{img}}[:, i, :].
    \end{equation}
    The image embedding is then routed to the corresponding scene based on cosine similarity:
    \begin{equation}
        \text{scene} = \arg\max \left( \cos(v_\text{e}, e_{\text{img}}) + b \right),
    \end{equation}
    where $v_\text{e}$ denotes the image embeddings $v$ projected into the external embedding space via $W_{\text{v}\rightarrow\text{e}} \in \mathbb{R}^{d_\text{v} \times d_\text{e}}$, and $b \in\mathbb{R}^{N_\text{s}}$ is a per-scene router balancing bias adaptively updated from the discrepancy between empirical routing frequencies and predicted probabilities~\cite{deepseekai2024deepseekv3technicalreport}. During training, the non-differentiable argmax is relaxed via a Gumbel-Softmax straight-through estimator.   

    \item \textbf{Level-2: scene-primitive retrieval.}
     Level-1 routes embeddings into coarse scenes, while level-2 refines details by generating soft weights over scene primitives to synthesize context-relevant embeddings. Level-2 cannot employ the same routing mechanism as level-1. While the routing embedding $e_{\text{img}}$ in level-1 is fixed and loaded only once, tokens in level-2 may route to different scenes, requiring frequent loading of all scene primitives and resulting in an MoE-like trade-off between loading latency and memory overhead. Since all scenes share the same primitive structure, we use a scene-shared two-layer MLP to estimate primitive relevance. For a level-1 scene $j$, relevance logit $r_j\in\mathbb{R}^{N_\text{p}}$ is computed by concatenating the projected image embedding $v_e$ and its scene-centered embedding $e_{\text{img}}[j, :]$, and the top-$\tilde{T}$ primitives $\mathcal{I}_{\tilde{T}}$ are retrieved as: 
    \begin{equation}
        r_j = \text{MLP}([v_\text{e}, e_{\text{img}}[j, :]]), \quad \mathcal{I}_{\tilde{T}} = \arg \text{Top}\tilde{T} \{r_j\}.
    \end{equation}
    The estimated primitives are further refined by relevance logits and cosine similarity to accommodate diverse inputs, then integrated into the external image embeddings:
    \begin{equation}
        \mathcal{R}_{\text{img}}  = \sum_{p \in \mathcal{I}_{\tilde{T}}} \phi
     \left( \cos(v_\text{e}, e_{j,p}) + \tau(r_j[p]) \right) \cdot e_{j,p},
    \end{equation}
    where $\phi$ and $\tau$ are softmax and tanh functions, $e_{j,p}$ denotes  $E_{\text{img}}[j,p,:] \in \mathbb{R}^{d_\text{e}}$, representing the embedding of the $p$-th primitive in scene $j$. During training, all external embeddings are loaded, while inference loads only a small number of estimated primitives for each token.
    
    \item \textbf{Multi-scale image retrieval.}
    To capture visual features at different granularities, we perform independent hierarchical two-level retrieval on image embeddings from $M$ multiple intermediate layers of the visual encoder $\{v_m\}_{m=1}^{M}$, each with its own coarse ranker and a table of fixed total capacity evenly partitioned across scenes by scale. The retrieved multi-scale embeddings are then fused at each image position via a token-wise softmax router conditioned on hidden states, enabling adaptive granularity selection for each token:
    \begin{equation}
    \begin{gathered}
        \mathcal{R}_{\text{img}} = \phi \left( (H_{\text{img}} W_{\text{h}\rightarrow\text{e}} \cdot \tilde{\mathcal{R}}_\text{img}^\top) / \sqrt{d_\text{e}} \right) \tilde{\mathcal{R}}_\text{img},  \\
      \quad \tilde{\mathcal{R}}_\text{img} = [\mathcal{R}^{1}_{\text{img}}, \dots, \mathcal{R}^{m}_{\text{img}}, \dots, \mathcal{R}^{M}_{\text{img}}],
    \end{gathered}
    \end{equation}
    where $W_{\text{h}\rightarrow\text{e}} \in \mathbb{R}^{d_\text{h} \times d_\text{e}}$ is a projection matrix, $H_{\text{img}}$ denotes the image-position hidden states of $\mathcal{L}_{\text{dec}}(H_{\text{in}})$, $\mathcal{R}^{m}_{\text{img}}$ represents the retrieved embeddings using the image embedding at scale $m$. 
\end{itemize}

\subsubsection{Cross-Modal Embedding Hierarchical Two-Level Lookup}
The cross-modal pathway operates on all positions. The retrieval input $X_\text{cross}$ is $\mathcal{L}_{\text{dec}}(H_{\text{in}})$ in each LookME layer, as they directly provide the cross-modal information. The cross-modal pathway also uses the hierarchical two-level lookup, independently from the image pathway, to retrieve continuous external embeddings. Unlike modality-specific pathways, it enables flexible embedding integration across text, image, audio, and mixed modalities. Notably, this process differs from MLP-based embedding adapters, where each token is densely computed over the entire adapter. Scaling the adapter for better performance requires loading all parameters, effectively degenerating into full-online similarity matching, whereas our pathway retrieves the corresponding enhancement information offline under sparse activation.

\subsection{Sparse Injection}
\label{sec:sparse-injection}
For a given LookME layer $\mathcal{L}_{\text{aug}}^{(l)}$,  the multimodal embedding lookup module retrieves three embeddings $\{\mathcal{R}_{\text{text}}^{(l)}, \mathcal{R}_{\text{img}}^{(l)}, \mathcal{R}_{\text{cross}}^{(l)}\}$, which are subsequently integrated through the injection function $\mathcal{S}$, as defined in Equation~\ref{eq:layer-output}. However, both retrieval and injection introduce additional overhead, particularly for the image and cross-modal pathways with long sequences, while converting every decoder layer into a LookME layer greatly increases model size. To address this, we propose a sparse injection method. The intra-layer sparse activation module filters low-signal tokens for efficient retrieval and injection, while the inter-layer sparse activation module activates LookME only in selected layers and propagates retrieved embeddings across non-LookME layers, avoiding repeated retrieval while preserving shallow-layer knowledge. Details are described below.

\subsubsection{Intra-Layer Sparse Activation}
\label{sec:ska}
The intra-layer sparse activation is controlled by sparse knowledge activation (SKA), which is a per-token, per-pathway binary gate to decide whether to invoke the embedding pathway at position $i$. SKA gate is computed as:
\begin{equation}
    \begin{gathered}
        g_{i} = \sigma(\Delta_iW_{\text{g}}),  \quad \Delta_i = \mathrm{Attn}(H_{\text{in}})_i,
    \end{gathered}
\end{equation}
where $W_{\text{g}} \in \mathbb{R}^{d_\text{h} \times 1} $ denotes the gate projection, $\sigma$ represents the Gumbel-Sigmoid function, which remains soft during training for differentiability and exploration while reducing to hard during inference. $\Delta_i$ denotes the attention contribution of the host LookME layer, specifically the additive update produced by self-attention at position $i$ before the residual connection in decoder layer. $\Delta_i$ is obtained at no additional computational cost, and tokens with large $|\Delta_i|$ are substantially updated by self-attention and are therefore more likely to benefit from external embedding augmentation, whereas tokens with near-zero values remain largely unchanged and receive limited benefit from additional embeddings. By conditioning on the shared attention contribution while applying distinct gate projections for multimodal pathways, the SKA gate is able to adaptively learn suitable pathway-specific activation patterns.

Each pathway in the multimodal embedding lookup module is executed only on the activated positions:
\begin{equation}
    \mathcal{S}(\mathcal{R_{*}}) = g_{*} \odot f_{*}\bigl(\mathcal{R_{*}}[g_{*}]\bigr),
\end{equation}
where $f$ denotes a SwiGLU-style FFN fusion that projects the retrieved embeddings from dimension ${d_{\text{e}}}$ to the backbone dimension $d_{\text{h}}$, and $\mathcal{S}(\mathcal{R}_{*})$ is further fused with the backbone sequentially as Equation~\eqref{eq:layer-output}.

\subsubsection{Inter-Layer Sparse Activation}
\label{sec:intra-layer}
Inter-layer sparse activation selectively transforms only a subset of decoder layers into LookME layers, thereby avoiding the substantial parameter overhead of a full conversion while still allowing non-LookME layers to benefit from the enhancement. Considering that adjacent layers tend to retrieve similar scene primitives, we convert only a small subset of layers into LookME layers, while allowing each intermediate non-LookME layer to inherit the sparse injection output from the preceding LookME layer through a lightweight bottleneck propagation, instead of re-running the full retrieval process. 

Assuming $k$ is a non-LookME layer and $l$ is the nearest LookME layer above $k$, the sparse activation for the layer $k$ reuses SKA gate from layer $l$ and performs residual propagation as: 
\begin{equation}
    \begin{gathered}
    \mathcal{S}^{(k)}(\mathcal{R_{*}}^{(k)}) = g^{(l)} \odot f^{k}\bigl(\mathcal{R_{*}}^{(k)}
    [g^{(l)}]\bigr), \\
    \mathcal{R_{*}}^{(k)} = \mathcal{R_{*}}^{(k-1)}W_\text{p}^{k} + \mathcal{R_{*}}^{(k-1)}, \quad k \geq l+1,
    \end{gathered}
\end{equation}
where $W_\text{p}^{k} \in \mathbb{R}^{d_{\text{e}} \times d_{\text{e}}}$ denotes the propagation projection in the embedding space, which is more lightweight than the backbone space since $d_{\text{e}} \ll d_{\text{h}}$. Through inter-layer sparse activation, enhancements to the backbone network become significantly more flexible and adaptable.

\section{Experiments}
In this section, we evaluate LookME against state-of-the-art methods, reporting the main results and ablation studies. Additional experiments, including analyses of hyperparameters, external visual embeddings, SKA activation patterns, and more, are provided in the supplementary material.

\subsection{Experimental Setup}
\subsubsection{Training Details}
Our training corpus is constructed entirely from open-source image--text datasets, including but not limited to~\cite{gu2024infinitymmscalingmultimodalperformance,wiedmann2025finevisionopendataneed,tong2024cambrian1fullyopenvisioncentric,yuan2025tarsier2advancinglargevisionlanguage,chen2025expandingperformanceboundariesopensource,chen2024allava,deitke2024molmopixmoopenweights}. We curate and sample data from representative vision tasks to ensure a balanced capability distribution. The final dataset contains 60M samples, comprising approximately 57B visual tokens and 14B text tokens. For fair comparison, all models are pre-trained on the same unified dataset. We adopt Qwen25VL-3B~\cite{bai2025qwen25vltechnicalreport} as the backbone for its strong multimodal capability and efficiency on edge devices. We perform full-parameter fine-tuning to LookME and all text-only PLE baselines using same external embedding layers 2/10/18 with dimension 512, following the official settings otherwise. For LookME, the $N_\text{s}$, $N_\text{p}$, top-$\tilde{T}$ are set to 512, 256, and 8, respectively, and the last three visual encoder layers 29/30/31 are used for multiscale representation. Detailed training datasets, configurations, environments, and hyperparameter analyses are provided in the supplementary material.

\subsubsection{Evaluation Details}
We compare LookME against three categories of baselines. First, the original \emph{Qwen25-VL} serves as a general baseline. Second, we include a baseline continued pre-trained on our unified dataset to isolate the gains brought by the training data itself, denoted as \emph{+CPT}. Third, since there are no corresponding multimodal PLE-style baselines, we compare against state-of-the-art text-only PLE-style methods, including \emph{PLE}~\cite{google2025gemma3n}, \emph{Engram}~\cite{cheng2026conditionalmemoryscalablelookup} and \emph{MeKi}~\cite{ding2026mekimemorybasedexpertknowledge}, to validate the effectiveness of our multimodal embedding lookup. To comprehensively evaluate performance, we conduct experiments on 15 established benchmarks covering diverse visual tasks, including \emph{STEM}: MMMU~\cite{yue2024mmmumassivemultidisciplinemultimodal}, MathVision~\cite{wang2024measuring}, and LogicVista~\cite{xiao2024logicvistamultimodalllmlogical}; \emph{General VQA}: MMBench-EN~\cite{liu2024mmbench}, MMStar~\cite{chen2024rightwayevaluatinglarge}, and RealWorldQA~\cite{realworldqa2024}; \emph{Hallucination}: HallusionBench~\cite{Guan_2024_CVPR}; \emph{Document Understanding}: InfoVQA~\cite{mathew2021infographicvqa} and OCRBench~\cite{Liu_2024}; \emph{Visual Perception}: HRBench8K~\cite{wang2024divideconquercombinetrainingfree}, CV-Bench-3D~\cite{tong2024cambrian1fullyopenvisioncentric}, and CCBench~\cite{liu2024mmbench}; \emph{Multi-Image}: BLINK~\cite{fu2024blinkmultimodallargelanguage}; \emph{Video}: Video-MME~\cite{fu2025video} and LongVideoBench~\cite{wu2024longvideobench}.

\subsection{Main Results}
\subsubsection{Comparison with SOTA Methods}
\begin{table*}[t]
    \centering
    \small
    \renewcommand{\arraystretch}{1.18}

    \begin{tabular}{clcccccc}

        \toprule

        \multirow{2}{*}{\textbf{Task}} &
        \multirow{2}{*}{\textbf{Benchmark}} &
        \multicolumn{6}{c}{\textbf{Models}} \\

        \cmidrule{3-8}

        & &
        \textbf{Qwen25-VL} &
        \textbf{+CPT} &
        \textbf{PLE} &
        \textbf{Engram} &
        \textbf{MeKi} &
        \textbf{LookME} \\

        \midrule

        \multicolumn{2}{l}{\textit{LLM Params (Backbone / External)}} &
        3.09B /- &
        3.09B /- &
        3.09B / 0.24B &
        3.09B / 2.35B &
        3.09B / 0.26B &
        3.09B / 0.82B \\

        \midrule

        \multirow{3}{*}{\textbf{STEM}}
        & MMMU
        & 46.9
        & \underline{47.1}
        & 43.8
        & 41.8
        & 46.3
        & \textbf{48.7} \\

        & MathVision
        & 17.9
        & 18.1
        & 17.3
        & 17.4
        & \underline{18.2}
        & \textbf{20.4} \\

        & LogicVista
        & 34.7
        & 36.5
        & 35.8
        & 32.7
        & \textbf{36.9}
        & \textbf{36.9} \\

        \cmidrule{1-8}

        \multirow{3}{*}{\textbf{General VQA}}
        & MMBench-EN
        & 78.3
        & \underline{78.7}
        & 73.7
        & 73.7
        & 77.1
        & \textbf{79.8} \\

        & MMStar
        & 55.1
        & \underline{55.7}
        & 53.5
        & 54.2
        & 54.5
        & \textbf{57.3} \\

        & RealWorldQA 
        & 65.5
        & 66.0
        & 62.2
        & 64.9
        & \underline{66.9}
        & \textbf{67.6} \\

        \cmidrule{1-8}

        \textbf{Hallucination}
        & HallusionBench
        & 65.2
        & 65.8
        & 67.7
        & 68.0
        & \underline{69.2}
        & \textbf{70.1} \\

        \cmidrule{1-8}

        \multirow{2}{*}{\textbf{\shortstack[c]{Document\\Understanding}}}
        & InfoVQA
        & 75.5
        & \underline{76.1}
        & 68.0
        & 69.1
        & 74.4
        & \textbf{77.2} \\

        & OCRBench
        & 82.9
        & \underline{83.7}
        & 82.1
        & 82.5
        & 82.7
        & \textbf{85.1} \\

        \cmidrule{1-8}

        \multirow{3}{*}{\textbf{\shortstack[c]{Visual\\Perception}}}
        & HRBench8K
        & 63.3
        & \underline{63.4}
        & 61.7
        & 60.9
        & 62.9
        & \textbf{66.5} \\

        & CV-Bench-3D
        & 74.8
        & \underline{75.3}
        & 73.4
        & 71.3
        & 73.6
        & \textbf{79.2} \\

        & CCBench
        & 62.6
        & \underline{63.7}
        & 62.3
        & 61.6
        & \underline{63.7}
        & \textbf{66.1} \\

        \cmidrule{1-8}

        \textbf{Multi-Image}
        & BLINK
        & 48.6
        & 47.6
        & 46.7
        & 47.8
        & \textbf{49.0}
        & \underline{48.7} \\

        \cmidrule{1-8}

        \multirow{2}{*}{\textbf{Video}}
        & Video-MME
        & 59.2
        & 59.8
        & 57.7
        & 58.3
        & \underline{60.4}
        & \textbf{61.4} \\

        & LongVideoBench
        & 54.7
        & \underline{54.9}
        & 51.3
        & 52.6
        & 54.5
        & \textbf{55.5} \\

        \bottomrule

    \end{tabular}
    \caption{
        Comparison with SOTA methods.
        The best results are in \textbf{bold} and the second best are
        \underline{underlined}. All models are evaluated under the same zero-shot settings.
    }
    \label{tab:sota}
\end{table*}

\label{sec:sota}
Table~\ref{tab:sota} reports results on 15 benchmarks across 7 task categories. LookME achieves the best performance on most datasets, and its advantage becomes more pronounced on tasks requiring fine-grained multimodal alignment. The gain is modest on standard single-image benchmarks (e.g., \textbf{+1.6} on MMMU), but larger on challenging visual perception benchmarks (\textbf{+2.8}--\textbf{+3.9}, e.g., CV-Bench-3D: \textbf{79.2} vs.\ \underline{75.3}). Notably, text-only PLE-style methods such as PLE, Engram, and MeKi underperform the backbone on most benchmarks. We attribute these results to LookME’s multimodal design. Unlike text-only PLE-style methods, LookME models image and cross-modal interactions and integrates retrieved information into the visual-language space. This is especially effective for evidence aggregation and multi-image alignment, where LookME achieves its largest gains.

\subsubsection{Analysis of Hierarchical Two-Level Lookup}
\begin{figure}[t]
\centering
\includegraphics[width=0.95\columnwidth]{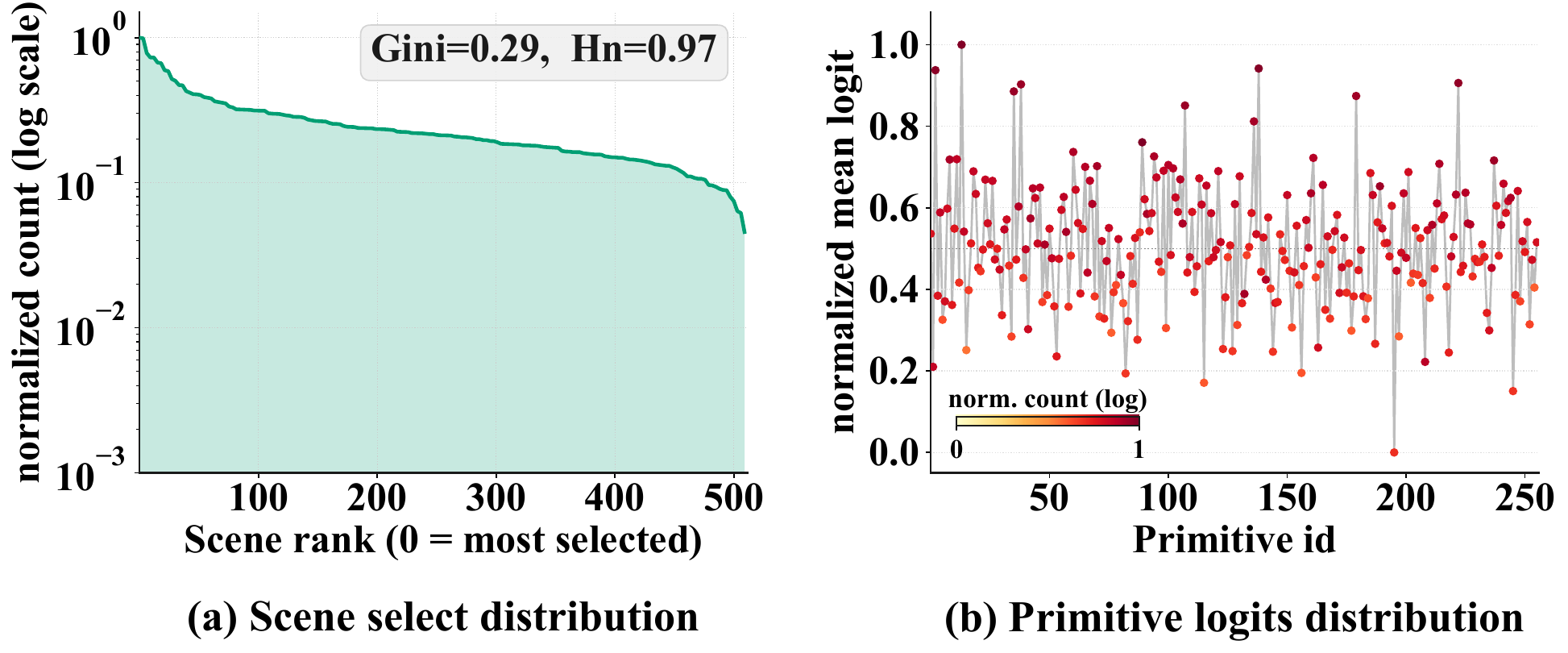} 
\caption{Statistical analysis of scene and primitive patterns.}
\label{fig:fig_scene_prim}
\end{figure}
The effectiveness of external embeddings has been validated in terms of accuracy. To further understand their behavior, we analyze the image external embeddings on RealWorldQA as a representative example. Figure~\ref{fig:fig_scene_prim}(a) shows the scene distribution in the level-1 lookup stage. All scenes can be selected, indicating that the retrieval process does not collapse, while the long-tail distribution suggests that the lookup is meaningful rather than random. Figure~\ref{fig:fig_scene_prim}(b) presents the primitive distribution in the level-2 lookup stage. For each primitive position, we compute the mean relevance logits under different selected scenes and visualize the normalized selection frequency, where darker dots indicate higher selection frequency. The results show that primitives at all positions can be activated, demonstrating effective retrieval. Moreover, the primitives exhibit both statistical and diversity properties. We found that the selection probabilities of different primitives vary, indicating an inherent combinatorial property rather than an imbalanced routing process. A small subset of primitives is frequently sampled in partial form and dynamically adjusted according to input relevance, potentially serving as shared statistical components for image representation. In contrast, a larger subset is sparsely sampled and dynamically combined, potentially capturing the distinctive characteristics of each image embedding. As a result, although image embeddings are highly diverse, LookME can still construct effective corresponding external embeddings for each one.

\subsubsection{Efficiency Analysis}
\label{sec:efficiency}
\begin{table}[t]
  \centering
  \small
  \begin{tabular*}{\columnwidth}{@{}l@{\extracolsep{\fill}}rrrr@{}}
  \toprule
  \textbf{Method} & \textbf{Mem. (GB)} & \textbf{Lat. (ms)} & \textbf{Thr. (t/s)} & \textbf{FLO. (G)} \\
  \midrule
  Backbone        & 7.74 & $\sim$160 & 2,021 & 5,168 \\
1-L (sim)  & +2.37 & +91 & -528 & +852 \\
1-L (cls)  & +4.11 & +93 & -559 & +2,065 \\
w/o Inter  & +0.79 & +25 & -139 & +331 \\
w/o Intra  & +0.83 & +39 & -258 & +736 \\
\textbf{LookME} & +0.83 & +30 & -203 & +352 \\
  \bottomrule
  \end{tabular*}
  \caption{Efficiency comparison. Mem. (Peak memory, GB), Lat. (latency, ms), Thr. (throughput, token/s), Flo. (FLOPs, G).}
  \label{tab:efficiency}
\end{table}

We analyze the efficiency under the training environment mentioned in the supplementary material. As shown in Table~\ref{tab:efficiency}, LookME adds retrieval and enhancement modules on top of the backbone, yet the overall overhead remains modest in memory, latency, and FLOPs, while still preserving efficient inference and improving performance. Compared with the two-level LookME, the one-level 1-L(sim) and 1-L(cls) in related work directly retrieve from the full vocabulary using similarity-based and classification-based strategies, respectively. Under the same embedding size, both variants are consistently less efficient, as in 1-L(sim), the entire embedding table must be loaded and computed for each token, while in 1-L(cls), the classification head grows linearly with vocabulary size. LookME uses a shard-compacted scene-center embedding and a shared MLP at each hierarchical level to restrict retrieval to smaller candidate sets, improving efficiency. Removing intra-layer sparse activation increases computational cost and reduces throughput, as all tokens are forced to participate in retrieval augmentation. This indicates that content-aware routing effectively avoids unnecessary retrieval. In contrast, inter-layer sparse activation adds only marginal overhead, while the connection enables lightweight global information propagation. Overall, the two-level retrieval and sparse activation mechanisms in LookME achieve a strong accuracy–efficiency trade-off with only moderate computational overhead.

\subsection{Ablation Experiment}
\subsubsection{Effectiveness of Multimodal Embedding Lookup}
\begin{table}[t]
\centering
\small
\begin{tabular*}{\columnwidth}{@{\extracolsep{\fill}}ccc|ccc c}
\toprule
\multicolumn{3}{c|}{\textbf{Modalities}} & \multicolumn{4}{c}{\textbf{Benchmarks}} \\
\cmidrule{1-3}\cmidrule{4-7}
\multicolumn{1}{c}{\textit{T}} & \multicolumn{1}{c}{\textit{I}} & \multicolumn{1}{c|}{\textit{C}} & \multicolumn{1}{c}{\textbf{MMMU}} & \multicolumn{1}{c}{\textbf{MMB.}} & \multicolumn{1}{c}{\textbf{OCR.}} & \multicolumn{1}{c}{\textbf{HRB.}} \\
\midrule
\multicolumn{1}{c}{$\checkmark$} & \multicolumn{1}{c}{$\times$} & \multicolumn{1}{c|}{$\times$} & \multicolumn{1}{c}{47.1} & \multicolumn{1}{c}{78.2} & \multicolumn{1}{c}{83.7} & \multicolumn{1}{c}{63.9} \\
\multicolumn{1}{c}{$\checkmark$} & \multicolumn{1}{c}{$\checkmark$} & \multicolumn{1}{c|}{$\times$} & \multicolumn{1}{c}{47.7} & \multicolumn{1}{c}{79.3} & \multicolumn{1}{c}{84.6} & \multicolumn{1}{c}{65.8} \\
\multicolumn{1}{c}{$\checkmark$} & \multicolumn{1}{c}{$\checkmark$} & \multicolumn{1}{c|}{$\checkmark$} & \multicolumn{1}{c}{\textbf{48.7}} & \multicolumn{1}{c}{\textbf{79.8}} & \multicolumn{1}{c}{\textbf{85.1}} & \multicolumn{1}{c}{\textbf{66.5}} \\
\bottomrule
\end{tabular*}
\caption{Ablation on multimodal lookup pathways. $\checkmark$ / $\times$ indicate whether the text $T$, image $I$, and cross-modal $C$ pathways are enabled or disabled. MMB.: MMBench-EN, OCR.: OCRBench, HRB.: HRBench8K. Complete results are in the supplementary material.}
\label{tab:ablation_multiembedding}
\end{table}

LookME extends embedding lookup from the prior text-only pathway to jointly constructed text, image, and cross-modal pathways. To study the contribution of each pathway, we progressively expand the retrieval input from \emph{Text-only} to \emph{Text+Img}, and finally to the full \emph{Text+Img+Cross} setting. As shown in Table~\ref{tab:ablation_multiembedding}, performance consistently improves as additional modalities are introduced. Incorporating image features already yields noticeable gains across all benchmarks, while the full multimodal configuration achieves the best overall results. These findings suggest that different modalities provide complementary retrieval signals, and demonstrate the importance of aligning retrieval with the backbone’s joint visual-textual representation space.

\subsubsection{Effectiveness of Sparse Injection}
LookME introduces sparsity at two levels: intra-layer sparse activation through \emph{SKA}, and inter-layer sparse activation through lightweight net propagation \emph{NetProp}. We analyze the two components separately in Table~\ref{tab:ablation_sparse}.

\begin{table}[t]
  \centering
  \small
  \begin{tabular*}{0.99\columnwidth}{@{\extracolsep{\fill}}c|cccc}
  \toprule
  \textbf{Variant} & \textbf{MMMU} & \textbf{MMB.} & \textbf{OCRB.} & \textbf{HRB.} \\
  \midrule
  \multicolumn{5}{l}{\textit{Intra-layer Sparse Activate (SKA)} \hfill \emph{default: Sparse}} \\
  \cmidrule(lr){1-5}
  Dense  & 48.6          & 79.4          & 84.9          & \textbf{67.5} \\
  Sparse & \textbf{48.7} & \textbf{79.8} & \textbf{85.1} & 66.5 \\
  \midrule
  \multicolumn{5}{l}{\textit{Inter-layer Sparse Activate (NetProp)} \hfill \emph{default: Head-Tail}} \\
  \cmidrule(lr){1-5}
  None      & 47.0          & 79.7          & 84.9          & 65.0 \\
  Full      & 47.8          & 78.8          & 84.3          & \textbf{67.6} \\
  Head-Tail & \textbf{48.7} & \textbf{79.8} & \textbf{85.1} & 66.5 \\
  \bottomrule
  \end{tabular*}
  \caption{Ablation on sparse injection: \emph{intra-layer sparse activation} (SKA) and \emph{inter-layer sparse activation} (NetProp). Complete results are in the supplementary material.}
  \label{tab:ablation_sparse}
\end{table}

\begin{itemize}
\item \textbf{Intra-layer: SKA.} Instead of activating all tokens uniformly, SKA dynamically selects informative tokens for LookME augmentation. Interestingly, the upper panel of Table~\ref{tab:ablation_sparse} shows that the sparse setting performs nearly identically to the dense counterpart, and even slightly better on several benchmarks. This indicates that the benefits of retrieval augmentation are concentrated on a relatively small subset of tokens, allowing SKA to reduce computation substantially without sacrificing representation quality.

\item \textbf{Inter-layer: NetProp.} 
We compare three propagation strategies: \emph{None} disables propagation, \emph{Full} propagates from the first LookME layer to the last LLM layer, and \emph{Head-Tail} propagates between the first and last LookME layers. Removing NetProp consistently degrades performance, confirming the need for cross-layer communication. While \emph{Full} improves HRBench8K, it hurts MMBench-EN and OCRBench. In contrast, \emph{Head-Tail} achieves the most balanced results, suggesting moderate propagation is sufficient and avoids excessive feature drift.
\end{itemize}

\subsubsection{Limitation}
LookME is continually pre-trained from a foundation model. We believe that training it from scratch on more comprehensive and diverse multimodal data could further enhance knowledge retention and memory capabilities, ultimately leading to improved performance.

\section{Conclusions}
In this paper, we proposed LookME, the first method that extends PLE lookup-based enhancement from text to multimodal embeddings in VLMs, addressing the limitation that existing PLE-style methods only enhance text tokens while leaving dominant multimodal information untouched. LookME introduces a hierarchical two-level lookup method that first coarsely routes continuous multimodal embeddings to scenes, then performs fine-grained retrieval via soft combination of scene-primitives, enabling efficient and balanced lookup at text-equivalent vocabulary scales. A sparse injection strategy further adaptively activates lookup based on attention signals and propagates retrieved embeddings across layers through lightweight transitions, achieving a favorable trade-off among efficiency, model size, and performance. Extensive experiments demonstrate that LookME outperforms baselines and text-only PLE methods across diverse visual benchmarks. LookME supports offline table storage and flexible pathway configuration, offering a promising direction for parameter-efficient multimodal model expansion.

\bibliography{aaai2027}

\begin{thebibliography}{46}
\providecommand{\natexlab}[1]{#1}

\bibitem[{Bai et~al.(2025)Bai, Chen, Liu, Wang, Ge, Song, Dang, Wang, Wang, Tang, Zhong, Zhu, Yang, Li, Wan, Wang, Ding, Fu, Xu, Ye, Zhang, Xie, Cheng, Zhang, Yang, Xu, and Lin}]{bai2025qwen25vltechnicalreport}
Bai, S.; Chen, K.; Liu, X.; Wang, J.; Ge, W.; Song, S.; Dang, K.; Wang, P.; Wang, S.; Tang, J.; Zhong, H.; Zhu, Y.; Yang, M.; Li, Z.; Wan, J.; Wang, P.; Ding, W.; Fu, Z.; Xu, Y.; Ye, J.; Zhang, X.; Xie, T.; Cheng, Z.; Zhang, H.; Yang, Z.; Xu, H.; and Lin, J. 2025.
\newblock Qwen2.5-VL Technical Report.
\newblock arXiv:2502.13923.

\bibitem[{Berges et~al.(2025)Berges, O{\u{g}}uz, Haziza, Yih, Zettlemoyer, and Ghosh}]{berges2024memorylayers}
Berges, V.-P.; O{\u{g}}uz, B.; Haziza, D.; Yih, W.-t.; Zettlemoyer, L.; and Ghosh, G. 2025.
\newblock Memory Layers at Scale.
\newblock In \emph{Proceedings of the International Conference on Machine Learning (ICML)}.

\bibitem[{Chen et~al.(2024{\natexlab{a}})Chen, Chen, Zhang, Chen, Wu, Zhang, Chen, Li, Wan, and Wang}]{chen2024allava}
Chen, G.~H.; Chen, S.; Zhang, R.; Chen, J.; Wu, X.; Zhang, Z.; Chen, Z.; Li, J.; Wan, X.; and Wang, B. 2024{\natexlab{a}}.
\newblock ALLaVA: Harnessing GPT4V-synthesized Data for A Lite Vision-Language Model.
\newblock arXiv:2402.11684.

\bibitem[{Chen et~al.(2024{\natexlab{b}})Chen, Li, Dong, Zhang, Zang, Chen, Duan, Wang, Qiao, Lin, and Zhao}]{chen2024rightwayevaluatinglarge}
Chen, L.; Li, J.; Dong, X.; Zhang, P.; Zang, Y.; Chen, Z.; Duan, H.; Wang, J.; Qiao, Y.; Lin, D.; and Zhao, F. 2024{\natexlab{b}}.
\newblock Are We on the Right Way for Evaluating Large Vision-Language Models?
\newblock arXiv:2403.20330.

\bibitem[{Chen et~al.(2025)Chen, Wang, Cao, Liu, Gao, Cui, Zhu, Ye, Tian, Liu, Gu, Wang, Li, Ren, Chen, Luo, Wang, Jiang, Wang, He, Shi, Zhang, Lv, Wang, Shao, Chu, Tu, He, Wu, Deng, Ge, Chen, Zhang, Wang, Dou, Lu, Zhu, Lu, Lin, Qiao, Dai, and Wang}]{chen2025expandingperformanceboundariesopensource}
Chen, Z.; Wang, W.; Cao, Y.; Liu, Y.; Gao, Z.; Cui, E.; Zhu, J.; Ye, S.; Tian, H.; Liu, Z.; Gu, L.; Wang, X.; Li, Q.; Ren, Y.; Chen, Z.; Luo, J.; Wang, J.; Jiang, T.; Wang, B.; He, C.; Shi, B.; Zhang, X.; Lv, H.; Wang, Y.; Shao, W.; Chu, P.; Tu, Z.; He, T.; Wu, Z.; Deng, H.; Ge, J.; Chen, K.; Zhang, K.; Wang, L.; Dou, M.; Lu, L.; Zhu, X.; Lu, T.; Lin, D.; Qiao, Y.; Dai, J.; and Wang, W. 2025.
\newblock Expanding Performance Boundaries of Open-Source Multimodal Models with Model, Data, and Test-Time Scaling.
\newblock arXiv:2412.05271.

\bibitem[{Chen et~al.(2024{\natexlab{c}})Chen, Wu, Wang, Su, Chen, Xing, Zhong, Zhang, Zhu, Lu, Li, Luo, Lu, Qiao, and Dai}]{chen2024internvl}
Chen, Z.; Wu, J.; Wang, W.; Su, W.; Chen, G.; Xing, S.; Zhong, M.; Zhang, Q.; Zhu, X.; Lu, L.; Li, B.; Luo, P.; Lu, T.; Qiao, Y.; and Dai, J. 2024{\natexlab{c}}.
\newblock InternVL: Scaling up Vision Foundation Models and Aligning for Generic Visual-Linguistic Tasks.
\newblock In \emph{Proceedings of the IEEE/CVF Conference on Computer Vision and Pattern Recognition (CVPR)}.
\newblock Oral.

\bibitem[{Cheng et~al.(2026)Cheng, Tian, Zeng, Dai, Chen, Wang, Xie, Huang, Yu, Deng, Zhou, Zhao, Hao, Li, Zhang, Zhang, Wei, Xu, Zhang, Zhao, and Liang}]{cheng2026conditionalmemoryscalablelookup}
Cheng, X.; Tian, R.; Zeng, W.; Dai, D.; Chen, Q.; Wang, B.; Xie, Z.; Huang, K.; Yu, X.; Deng, C.; Zhou, S.; Zhao, C.; Hao, Z.; Li, Y.; Zhang, H.; Zhang, Z.; Wei, Y.; Xu, M.~Y.; Zhang, H.; Zhao, D.; and Liang, W. 2026.
\newblock Conditional Memory via Scalable Lookup: A New Axis of Sparsity for Large Language Models.
\newblock arXiv:2601.07372.

\bibitem[{Dai et~al.(2024)Dai, Deng, Zhao, Xu, Gao, Chen, Li, Zeng, Yu, Wu, Xie, Li, Huang, Luo, Ruan, Sui, and Liang}]{dai2024deepseekmoeultimateexpertspecialization}
Dai, D.; Deng, C.; Zhao, C.; Xu, R.~X.; Gao, H.; Chen, D.; Li, J.; Zeng, W.; Yu, X.; Wu, Y.; Xie, Z.; Li, Y.~K.; Huang, P.; Luo, F.; Ruan, C.; Sui, Z.; and Liang, W. 2024.
\newblock DeepSeekMoE: Towards Ultimate Expert Specialization in Mixture-of-Experts Language Models.
\newblock arXiv:2401.06066.

\bibitem[{DeepSeek-AI(2024)}]{deepseekai2024deepseekv3technicalreport}
DeepSeek-AI. 2024.
\newblock DeepSeek-V3 Technical Report.
\newblock arXiv:2412.19437.

\bibitem[{Deitke et~al.(2024)Deitke, Clark, Lee, Tripathi, Yang, Park, Salehi, Muennighoff, Lo, Soldaini, Lu, Anderson, Bransom, Ehsani, Ngo, Chen, Patel, Yatskar, Callison-Burch, Head, Hendrix, Bastani, VanderBilt, Lambert, Chou, Chheda, Sparks, Skjonsberg, Schmitz, Sarnat, Bischoff, Walsh, Newell, Wolters, Gupta, Zeng, Borchardt, Groeneveld, Nam, Lebrecht, Wittlif, Schoenick, Michel, Krishna, Weihs, Smith, Hajishirzi, Girshick, Farhadi, and Kembhavi}]{deitke2024molmopixmoopenweights}
Deitke, M.; Clark, C.; Lee, S.; Tripathi, R.; Yang, Y.; Park, J.~S.; Salehi, M.; Muennighoff, N.; Lo, K.; Soldaini, L.; Lu, J.; Anderson, T.; Bransom, E.; Ehsani, K.; Ngo, H.; Chen, Y.; Patel, A.; Yatskar, M.; Callison-Burch, C.; Head, A.; Hendrix, R.; Bastani, F.; VanderBilt, E.; Lambert, N.; Chou, Y.; Chheda, A.; Sparks, J.; Skjonsberg, S.; Schmitz, M.; Sarnat, A.; Bischoff, B.; Walsh, P.; Newell, C.; Wolters, P.; Gupta, T.; Zeng, K.-H.; Borchardt, J.; Groeneveld, D.; Nam, C.; Lebrecht, S.; Wittlif, C.; Schoenick, C.; Michel, O.; Krishna, R.; Weihs, L.; Smith, N.~A.; Hajishirzi, H.; Girshick, R.; Farhadi, A.; and Kembhavi, A. 2024.
\newblock Molmo and PixMo: Open Weights and Open Data for State-of-the-Art Vision-Language Models.
\newblock arXiv:2409.17146.

\bibitem[{Ding et~al.(2026)Ding, Liu, Kim, Hao, Lee, Ko, and Tang}]{ding2026mekimemorybasedexpertknowledge}
Ding, N.; Liu, F.; Kim, K.; Hao, L.; Lee, K.-H.; Ko, H.; and Tang, Y. 2026.
\newblock MeKi: Memory-based Expert Knowledge Injection for Efficient LLM Scaling.
\newblock arXiv:2602.03359.

\bibitem[{Dosovitskiy et~al.(2021)Dosovitskiy, Beyer, Kolesnikov, Weissenborn, Zhai, Unterthiner, Dehghani, Minderer, Heigold, Gelly, Uszkoreit, and Houlsby}]{dosovitskiy2021vit}
Dosovitskiy, A.; Beyer, L.; Kolesnikov, A.; Weissenborn, D.; Zhai, X.; Unterthiner, T.; Dehghani, M.; Minderer, M.; Heigold, G.; Gelly, S.; Uszkoreit, J.; and Houlsby, N. 2021.
\newblock An Image is Worth 16x16 Words: Transformers for Image Recognition at Scale.
\newblock In \emph{Proceedings of the International Conference on Learning Representations (ICLR)}.

\bibitem[{Fedus, Zoph, and Shazeer(2022)}]{fedus2022switch}
Fedus, W.; Zoph, B.; and Shazeer, N. 2022.
\newblock Switch Transformers: Scaling to Trillion Parameter Models with Simple and Efficient Sparsity.
\newblock \emph{Journal of Machine Learning Research (JMLR)}, 23(120): 1--39.

\bibitem[{Fu et~al.(2025)Fu, Dai, Luo, Li, Ren, Zhang, Wang, Zhou, Shen, Zhang et~al.}]{fu2025video}
Fu, C.; Dai, Y.; Luo, Y.; Li, L.; Ren, S.; Zhang, R.; Wang, Z.; Zhou, C.; Shen, Y.; Zhang, M.; et~al. 2025.
\newblock Video-mme: The first-ever comprehensive evaluation benchmark of multi-modal llms in video analysis.
\newblock In \emph{Proceedings of the Computer Vision and Pattern Recognition Conference}, 24108--24118.

\bibitem[{Fu et~al.(2024)Fu, Hu, Li, Feng, Wang, Lin, Roth, Smith, Ma, and Krishna}]{fu2024blinkmultimodallargelanguage}
Fu, X.; Hu, Y.; Li, B.; Feng, Y.; Wang, H.; Lin, X.; Roth, D.; Smith, N.~A.; Ma, W.-C.; and Krishna, R. 2024.
\newblock BLINK: Multimodal Large Language Models Can See but Not Perceive.
\newblock arXiv:2404.12390.

\bibitem[{{Google DeepMind}(2025)}]{google2025gemma3n}
{Google DeepMind}. 2025.
\newblock Gemma 3n: On-Device Multimodal AI with Per-Layer Embeddings.
\newblock Technical report, Google DeepMind.

\bibitem[{Gu et~al.(2024)Gu, Zhang, Zhou, Yu, Xing, Wang, Cao, Jia, Zhang, Wang, Hu, Zhang, Li, Liang, Zhao, Ao, Liu, Feng, and Liu}]{gu2024infinitymmscalingmultimodalperformance}
Gu, S.; Zhang, J.; Zhou, S.; Yu, K.; Xing, Z.; Wang, L.; Cao, Z.; Jia, J.; Zhang, Z.; Wang, Y.; Hu, Z.; Zhang, B.-W.; Li, J.; Liang, D.; Zhao, Y.; Ao, Y.; Liu, Y.; Feng, F.; and Liu, G. 2024.
\newblock Infinity-MM: Scaling Multimodal Performance with Large-Scale and High-Quality Instruction Data.
\newblock arXiv:2410.18558.

\bibitem[{Guan et~al.(2024)Guan, Liu, Wu, Xian, Li, Liu, Wang, Chen, Huang, Yacoob, Manocha, and Zhou}]{Guan_2024_CVPR}
Guan, T.; Liu, F.; Wu, X.; Xian, R.; Li, Z.; Liu, X.; Wang, X.; Chen, L.; Huang, F.; Yacoob, Y.; Manocha, D.; and Zhou, T. 2024.
\newblock HallusionBench: An Advanced Diagnostic Suite for Entangled Language Hallucination and Visual Illusion in Large Vision-Language Models.
\newblock In \emph{Proceedings of the IEEE/CVF Conference on Computer Vision and Pattern Recognition (CVPR)}, 14375--14385.

\bibitem[{Gupta, Singaravadivelan, and Wang(2025)}]{gupta2025cobweb}
Gupta, A.; Singaravadivelan, K.; and Wang, Z. 2025.
\newblock Hierarchical Semantic Retrieval with Cobweb.
\newblock arXiv:2510.02539.

\bibitem[{Gupta et~al.(2025)Gupta, Li, Chen, Subakan, Reddy, Taslakian, and Zantedeschi}]{gupta2025retreever}
Gupta, S.; Li, Z.; Chen, T.; Subakan, C.; Reddy, S.; Taslakian, P.; and Zantedeschi, V. 2025.
\newblock Hierarchical Retrieval at Scale: Bridging Transparency and Efficiency.
\newblock arXiv:2502.07971.

\bibitem[{Hoffmann et~al.(2022)Hoffmann, Borgeaud, Mensch, Buchatskaya, Cai, Rutherford, de~Las~Casas, Hendricks, Welbl, Clark, Hennigan, Noland, Millican, van~den Driessche, Damoc, Guy, Osindero, Simonyan, Elsen, Rae, Vinyals, and Sifre}]{hoffmann2022chinchilla}
Hoffmann, J.; Borgeaud, S.; Mensch, A.; Buchatskaya, E.; Cai, T.; Rutherford, E.; de~Las~Casas, D.; Hendricks, L.~A.; Welbl, J.; Clark, A.; Hennigan, T.; Noland, E.; Millican, K.; van~den Driessche, G.; Damoc, B.; Guy, A.; Osindero, S.; Simonyan, K.; Elsen, E.; Rae, J.~W.; Vinyals, O.; and Sifre, L. 2022.
\newblock Training Compute-Optimal Large Language Models.
\newblock In \emph{Advances in Neural Information Processing Systems (NeurIPS)}.

\bibitem[{J{\'e}gou, Douze, and Schmid(2011)}]{jegou2011product}
J{\'e}gou, H.; Douze, M.; and Schmid, C. 2011.
\newblock Product Quantization for Nearest Neighbor Search.
\newblock \emph{IEEE Transactions on Pattern Analysis and Machine Intelligence (TPAMI)}, 33(1): 117--128.

\bibitem[{Kaplan et~al.(2020)Kaplan, McCandlish, Henighan, Brown, Chess, Child, Gray, Radford, Wu, and Amodei}]{kaplan2020scalinglawsneurallanguage}
Kaplan, J.; McCandlish, S.; Henighan, T.; Brown, T.~B.; Chess, B.; Child, R.; Gray, S.; Radford, A.; Wu, J.; and Amodei, D. 2020.
\newblock Scaling Laws for Neural Language Models.
\newblock arXiv:2001.08361.

\bibitem[{Khandelwal et~al.(2020)Khandelwal, Levy, Jurafsky, Zettlemoyer, and Lewis}]{khandelwal2020knnlm}
Khandelwal, U.; Levy, O.; Jurafsky, D.; Zettlemoyer, L.; and Lewis, M. 2020.
\newblock Generalization through Memorization: Nearest Neighbor Language Models.
\newblock In \emph{Proceedings of the International Conference on Learning Representations (ICLR)}.

\bibitem[{Lample et~al.(2019)Lample, Sablayrolles, Ranzato, Denoyer, and J{\'e}gou}]{lample2019productkey}
Lample, G.; Sablayrolles, A.; Ranzato, M.; Denoyer, L.; and J{\'e}gou, H. 2019.
\newblock Large Memory Layers with Product Keys.
\newblock In \emph{Advances in Neural Information Processing Systems (NeurIPS)}.

\bibitem[{Liu et~al.(2023)Liu, Li, Wu, and Lee}]{liu2023llava}
Liu, H.; Li, C.; Wu, Q.; and Lee, Y.~J. 2023.
\newblock Visual Instruction Tuning.
\newblock In \emph{Advances in Neural Information Processing Systems (NeurIPS)}.

\bibitem[{Liu et~al.(2026)Liu, Zhang, Wang, Hu, Lyu, Sun, Yang, Wang, Li, Qian, Si, Sun, Li, Pei, Xie, and Cai}]{liu2026scalingembeddingsoutperformsscaling}
Liu, H.; Zhang, J.; Wang, C.; Hu, X.; Lyu, L.; Sun, J.; Yang, X.; Wang, B.; Li, F.; Qian, Y.; Si, L.; Sun, Y.; Li, R.; Pei, P.; Xie, Y.; and Cai, X. 2026.
\newblock Scaling Embeddings Outperforms Scaling Experts in Language Models.
\newblock arXiv:2601.21204.

\bibitem[{Liu et~al.(2024{\natexlab{a}})Liu, Duan, Zhang, Li, Zhang, Zhao, Yuan, Wang, He, Liu et~al.}]{liu2024mmbench}
Liu, Y.; Duan, H.; Zhang, Y.; Li, B.; Zhang, S.; Zhao, W.; Yuan, Y.; Wang, J.; He, C.; Liu, Z.; et~al. 2024{\natexlab{a}}.
\newblock Mmbench: Is your multi-modal model an all-around player?
\newblock In \emph{European conference on computer vision}, 216--233. Springer.

\bibitem[{Liu et~al.(2024{\natexlab{b}})Liu, Li, Huang, Yang, Yu, Li, Yin, Liu, Jin, and Bai}]{Liu_2024}
Liu, Y.; Li, Z.; Huang, M.; Yang, B.; Yu, W.; Li, C.; Yin, X.-C.; Liu, C.-L.; Jin, L.; and Bai, X. 2024{\natexlab{b}}.
\newblock OCRBench: on the hidden mystery of OCR in large multimodal models.
\newblock \emph{Science China Information Sciences}, 67(12).

\bibitem[{Lu et~al.(2024{\natexlab{a}})Lu, Liu, Zhang, Wang, Dong, Liu, Sun, Ren, Li, Yang, Sun, Deng, Xu, Xie, and Ruan}]{lu2024deepseekvlrealworldvisionlanguageunderstanding}
Lu, H.; Liu, W.; Zhang, B.; Wang, B.; Dong, K.; Liu, B.; Sun, J.; Ren, T.; Li, Z.; Yang, H.; Sun, Y.; Deng, C.; Xu, H.; Xie, Z.; and Ruan, C. 2024{\natexlab{a}}.
\newblock DeepSeek-VL: Towards Real-World Vision-Language Understanding.
\newblock arXiv:2403.05525.

\bibitem[{Lu et~al.(2024{\natexlab{b}})Lu, Li, Chen, Xu, Luo, Zhang, and Ye}]{lu2024ovis}
Lu, S.; Li, Y.; Chen, Q.-G.; Xu, Z.; Luo, W.; Zhang, K.; and Ye, H.-J. 2024{\natexlab{b}}.
\newblock Ovis: Structural Embedding Alignment for Multimodal Large Language Model.
\newblock arXiv:2405.20797.

\bibitem[{Mathew et~al.(2021)Mathew, Bagal, Tito, Karatzas, Valveny, and Jawahar}]{mathew2021infographicvqa}
Mathew, M.; Bagal, V.; Tito, R.~P.; Karatzas, D.; Valveny, E.; and Jawahar, C.~V. 2021.
\newblock InfographicVQA.
\newblock arXiv:2104.12756.

\bibitem[{Sadhukhan et~al.(2026)Sadhukhan, Cao, Dong, Zhao, Purpura-Pontoniere, Tian, Liu, and Chen}]{stem2026}
Sadhukhan, R.; Cao, S.; Dong, H.; Zhao, C.; Purpura-Pontoniere, A.; Tian, Y.; Liu, Z.; and Chen, B. 2026.
\newblock STEM: Scaling Transformers with Embedding Modules.
\newblock In \emph{Proceedings of the International Conference on Learning Representations (ICLR)}.

\bibitem[{Shazeer et~al.(2017)Shazeer, Mirhoseini, Maziarz, Davis, Le, Hinton, and Dean}]{shazeer2017moe}
Shazeer, N.; Mirhoseini, A.; Maziarz, K.; Davis, A.; Le, Q.; Hinton, G.; and Dean, J. 2017.
\newblock Outrageously Large Neural Networks: The Sparsely-Gated Mixture-of-Experts Layer.
\newblock In \emph{Proceedings of the International Conference on Learning Representations (ICLR)}.

\bibitem[{Shen et~al.(2023)Shen, Yao, Li, Darrell, Keutzer, and He}]{shen2023vlmoe}
Shen, S.; Yao, Z.; Li, C.; Darrell, T.; Keutzer, K.; and He, Y. 2023.
\newblock Scaling Vision-Language Models with Sparse Mixture of Experts.
\newblock In \emph{Findings of the Association for Computational Linguistics: EMNLP}.

\bibitem[{Tong et~al.(2024)Tong, Brown, Wu, Woo, Middepogu, Akula, Yang, Yang, Iyer, Pan, Wang, Fergus, LeCun, and Xie}]{tong2024cambrian1fullyopenvisioncentric}
Tong, S.; Brown, E.; Wu, P.; Woo, S.; Middepogu, M.; Akula, S.~C.; Yang, J.; Yang, S.; Iyer, A.; Pan, X.; Wang, Z.; Fergus, R.; LeCun, Y.; and Xie, S. 2024.
\newblock Cambrian-1: A Fully Open, Vision-Centric Exploration of Multimodal LLMs.
\newblock arXiv:2406.16860.

\bibitem[{van~den Oord, Vinyals, and Kavukcuoglu(2017)}]{van2017vqvae}
van~den Oord, A.; Vinyals, O.; and Kavukcuoglu, K. 2017.
\newblock Neural Discrete Representation Learning.
\newblock In \emph{Advances in Neural Information Processing Systems (NeurIPS)}.

\bibitem[{Wang et~al.(2024{\natexlab{a}})Wang, Pan, Shi, Lu, Ren, Zhou, Zhan, and Li}]{wang2024measuring}
Wang, K.; Pan, J.; Shi, W.; Lu, Z.; Ren, H.; Zhou, A.; Zhan, M.; and Li, H. 2024{\natexlab{a}}.
\newblock Measuring Multimodal Mathematical Reasoning with MATH-Vision Dataset.
\newblock In \emph{The Thirty-eight Conference on Neural Information Processing Systems Datasets and Benchmarks Track}.

\bibitem[{Wang et~al.(2024{\natexlab{b}})Wang, Ding, Zeng, Zhou, Shen, Luo, and Tao}]{wang2024divideconquercombinetrainingfree}
Wang, W.; Ding, L.; Zeng, M.; Zhou, X.; Shen, L.; Luo, Y.; and Tao, D. 2024{\natexlab{b}}.
\newblock Divide, Conquer and Combine: A Training-Free Framework for High-Resolution Image Perception in Multimodal Large Language Models.
\newblock arXiv:2408.15556.

\bibitem[{Wiedmann et~al.(2025)Wiedmann, Zohar, Mahla, Wang, Li, Frere, von Werra, Gosthipaty, and Marafioti}]{wiedmann2025finevisionopendataneed}
Wiedmann, L.; Zohar, O.; Mahla, A.; Wang, X.; Li, R.; Frere, T.; von Werra, L.; Gosthipaty, A.~R.; and Marafioti, A. 2025.
\newblock FineVision: Open Data Is All You Need.
\newblock arXiv:2510.17269.

\bibitem[{Wu et~al.(2024{\natexlab{a}})Wu, Li, Chen, and Li}]{wu2024longvideobench}
Wu, H.; Li, D.; Chen, B.; and Li, J. 2024{\natexlab{a}}.
\newblock Longvideobench: A benchmark for long-context interleaved video-language understanding.
\newblock \emph{Advances in Neural Information Processing Systems}, 37: 28828--28857.

\bibitem[{Wu et~al.(2024{\natexlab{b}})Wu, Chen, Pan, Liu, Liu, Dai, Gao, Ma, Wu, Wang, Xie, Wu, Hu, Wang, Sun, Li, Piao, Guan, Liu, Xie, You, Dong, Yu, Zhang, Zhao, Wang, and Ruan}]{wu2024deepseekvl2mixtureofexpertsvisionlanguagemodels}
Wu, Z.; Chen, X.; Pan, Z.; Liu, X.; Liu, W.; Dai, D.; Gao, H.; Ma, Y.; Wu, C.; Wang, B.; Xie, Z.; Wu, Y.; Hu, K.; Wang, J.; Sun, Y.; Li, Y.; Piao, Y.; Guan, K.; Liu, A.; Xie, X.; You, Y.; Dong, K.; Yu, X.; Zhang, H.; Zhao, L.; Wang, Y.; and Ruan, C. 2024{\natexlab{b}}.
\newblock DeepSeek-VL2: Mixture-of-Experts Vision-Language Models for Advanced Multimodal Understanding.
\newblock arXiv:2412.10302.

\bibitem[{{xAI}(2024)}]{realworldqa2024}
{xAI}. 2024.
\newblock RealWorldQA: A Benchmark for Real-World Spatial Understanding.
\newblock \url{https://huggingface.co/datasets/xai-org/RealworldQA}.
\newblock Accessed: 2025-04-26.

\bibitem[{Xiao et~al.(2024)Xiao, Sun, Liu, and Wang}]{xiao2024logicvistamultimodalllmlogical}
Xiao, Y.; Sun, E.; Liu, T.; and Wang, W. 2024.
\newblock LogicVista: Multimodal LLM Logical Reasoning Benchmark in Visual Contexts.
\newblock arXiv:2407.04973.

\bibitem[{Yuan et~al.(2025)Yuan, Wang, Sun, Zhang, and Lin}]{yuan2025tarsier2advancinglargevisionlanguage}
Yuan, L.; Wang, J.; Sun, H.; Zhang, Y.; and Lin, Y. 2025.
\newblock Tarsier2: Advancing Large Vision-Language Models from Detailed Video Description to Comprehensive Video Understanding.
\newblock arXiv:2501.07888.

\bibitem[{Yue et~al.(2024)Yue, Ni, Zhang, Zheng, Liu, Zhang, Stevens, Jiang, Ren, Sun, Wei, Yu, Yuan, Sun, Yin, Zheng, Yang, Liu, Huang, Sun, Su, and Chen}]{yue2024mmmumassivemultidisciplinemultimodal}
Yue, X.; Ni, Y.; Zhang, K.; Zheng, T.; Liu, R.; Zhang, G.; Stevens, S.; Jiang, D.; Ren, W.; Sun, Y.; Wei, C.; Yu, B.; Yuan, R.; Sun, R.; Yin, M.; Zheng, B.; Yang, Z.; Liu, Y.; Huang, W.; Sun, H.; Su, Y.; and Chen, W. 2024.
\newblock MMMU: A Massive Multi-discipline Multimodal Understanding and Reasoning Benchmark for Expert AGI.
\newblock arXiv:2311.16502.

\end{thebibliography}


\end{document}